\documentclass[11pt,letterpaper]{article}
\pdfoutput=1
\usepackage{emnlp2017}
\usepackage{times}
\usepackage{latexsym}
\usepackage{amsmath}
\usepackage{amssymb}
\usepackage{url}
\usepackage{mathtools}
\usepackage{graphicx,accents}
\usepackage{bm}

\makeatletter
\DeclareRobustCommand{\cev}[1]{%
	\mathpalette\do@cev{#1}%
}

\newcommand{\do@cev}[2]{%
	\fix@cev{#1}{+}%
	\reflectbox{$\m@th#1\vec{\reflectbox{$\fix@cev{#1}{-}\m@th#1#2\fix@cev{#1}{+}$}}$}%
	\fix@cev{#1}{-}%
}

\newcommand{\fix@cev}[2]{%
	\ifx#1\displaystyle
	\mkern#23mu
	\else
	\ifx#1\textstyle
	\mkern#23mu
	\else
	\ifx#1\scriptstyle
	\mkern#22mu
	\else
	\mkern#22mu
	\fi
	\fi
	\fi
}
\makeatother

\emnlpfinalcopy

\def\emnlppaperid{***}

\title{Refining Raw Sentence Representations for Textual Entailment Recognition via Attention}

\author{Jorge A. Balazs, Edison Marrese-Taylor, Pablo Loyola, Yutaka Matsuo \\
  Graduate School of Engineering\\
  The University of Tokyo\\
  {\tt \{jorge,emarrese,pablo,matsuo\}@weblab.t.u-tokyo.ac.jp} \\}

\date{}

\begin{document}

\maketitle
\begin{abstract}
    In this paper we present the model used by the team Rivercorners for the 2017 RepEval shared task. First, our model separately encodes a pair of sentences into variable-length representations by using a bidirectional LSTM. Later, it creates fixed-length raw representations by means of simple aggregation functions, which are then refined using an attention mechanism. Finally it combines the refined representations of both sentences into a single vector to be used for classification. With this model we obtained test accuracies of $72.057\%$ and $72.055\%$ in the matched and mismatched evaluation tracks respectively, outperforming the LSTM baseline, and obtaining performances similar to a model that relies on shared information between sentences (ESIM). When using an ensemble both accuracies increased to $72.247\%$ and $72.827\%$ respectively.

\end{abstract}

\section{Introduction}


The task of Natural Language Inference (NLI)
aims at characterizing the semantic concepts of entailment and contradiction, and is essential in tasks ranging from information retrieval to semantic parsing to commonsense reasoning, as both entailment and contradiction are central concepts in natural language meaning \cite{katz1972semantic, van2008brief}.

The aforementioned task has been addressed with a variety of techniques, including those based on symbolic logic, knowledge bases, and neural networks. With the advent of deep learning techniques, NLI has become an important testing ground for approaches that employ distributed word and phrase representations, which are typical of these models.

In this context, the Second Workshop on Evaluating Vector Space Representations for NLP (RepEval 2017) features a shared task meant to evaluate natural language understanding models based on sentence encoders by the means of NLI in the style of a three-class balanced classification problem over sentence pairs. The shared task includes two evaluations, a standard in-domain (matched) evaluation in which the training and test data are drawn from the same sources, and a cross-domain (mismatched) evaluation in which the training and test data differ substantially. This cross-domain evaluation is aimed at testing the ability of submitted systems to learn representations of sentence meaning that capture broadly useful features.

\section{Proposed Model}

Our work is related to intra-sentence attention models for sentence representation such as the ones described by  \newcite{liu2017innerattention} and \newcite{lin2017selfattentive}. In particular, our model is based on the notion that, when reading a sentence, we usually need to re-read certain portions of it in order to  obtain a comprehensive understanding. To model such phenomenon, we rely on an attention mechanism able to iteratively obtain a richer and more expressive version of a raw sentence representation. The model's architecture is described below:



\textbf{Word Representation Layer}: This layer is in charge of generating a comprehensive vector representation of each token for a given sentence. We construct this representation based on up to two basic components:

\begin{itemize}
\item Pre-trained word embeddings: We take pre-trained word embeddings and use them to generate a raw word representation. This can be seen as a simple lookup-layer that returns a word vector for each provided word index.

\item Character embeddings: We generate a character-based representation of each word, which we concatenate to the word vectors as returned by the previous component. We start by generating a randomly initialized character embedding matrix $C$. Then, we split each word into its component characters, get their corresponding character embedding vectors from $C$ and feed them into a unidirectional Long Short-Term Memory Network (LSTM) \cite{hochreiter1997long}. We then choose the last hidden state returned by the LSTM as the fixed-size character-based vector representation for each token. Our embedding matrix $C$ is trained with the rest of the model \cite{wang2017multiperspective}.


\end{itemize}

\textbf{Context Representation Layer}: This layer complements the vectors generated by the Word Representation Layer by incorporating contextual information into them. To do this, we utilize a bidirectional LSTM that reads through the embedded sequence and returns the hidden states for each time step. These are context-aware representations focused on each position. Formally, let $\mathcal{S}$ be a sentence such as $\mathcal{S} = \{\bm{x}_1,\ldots,\bm{x}_n\}$, where each $\bm{x}_i$ is an embedded word vector as returned by the previous layer, then the context-rich word representation $\bm{h}_i$ is calculated as follows for each time step $i=1,\ldots,n$:
\begin{align}
&\overrightarrow{\bm{h}}_i = LSTM(\bm{x}_i, \overrightarrow{\bm{h}}_{i-1})\\
&\overleftarrow{\bm{h}}_i = LSTM(\bm{x}_i, \overleftarrow{\bm{h}}_{i+1})\\
&\bm{h}_i = [\overrightarrow{\bm{h}}_i ; \overleftarrow{\bm{h}}_i]
\end{align}


%

\noindent Where $\overrightarrow{\bm{h}}_i$ is the forward contextual vector representation of $\bm{x}_i$, $\overleftarrow{\bm{h}}_i$ the backward one, and $[\enskip \cdot \enskip;\enskip \cdot \enskip]$ represents the concatenation of two vectors. The output of this layer is a variable-length sentence representation for both the premise and hypothesis. We then define a pooling layer in charge of a generating a raw fixed-size representation of each sentence.

\textbf{Pooling Layer}: This layer is in charge of generating a crude sentence representation vector by reducing the sequence dimension using one of four simple operations, all of which are fed the context-aware token representations obtained previously:

\begin{align}
\label{mean-pool}\bar{\bm{h}} &= \frac1n \sum_{i=1}^n \bm{h}_i\\
\label{sum-pool}\bar{\bm{h}} &= \sum_{i=1}^n \bm{h}_i\\
\label{last-pool}\bar{\bm{h}} &= [\overrightarrow{\bm{h}}_n; \overleftarrow{\bm{h}}_1]\\
\label{max-pool}\bar{\bm{h}} &= \max_{i= 1 ... n} \bm{h}_i
\end{align}

\noindent These operations correspond to the \emph{mean} of the word representations (eq. \ref{mean-pool}), their \emph{sum} (eq. \ref{sum-pool}), the concatenation of the \emph{last} hidden state for each direction (eq. \ref{last-pool}), and the \emph{maximum} one (eq. \ref{max-pool}).

\textbf{Inner Attention Layer}: To refine the representations generated by the pooling strategy, we use a global attention mechanism \cite{luong_effective_2015, vinyals_grammar_2015} that compares each context-aware token representation $\bm{h}_i$ with the raw sentence representation $\bar{\bm{h}}$. Formally,
\begin{align}
	\label{att-scoring}u_i &= \bm{v}^{\top} \tanh(W [\bar{\bm{h}}; \bm{h}_i])\\
	\alpha_i &= \frac{\exp{u_i}}{\sum_{k=1}^{n}{\exp{u_k}}} \\
	\bar{\bm{h}}^{\prime} &= \sum_{i=1}^n \alpha_i \bm{h}_i
\end{align}

\noindent Where both $\bm{v}$ and $W$ are trainable parameters and $\bar{\bm{h}}^{\prime}$ is the refined sentence representation\footnote{The refined sentence representation $\bar{\bm{h}}^{\prime}$ for both premise and hypothesis is the final representation in which both are treated as separate entities. The representations produced by our best-performing model are available in \url{https://zenodo.org/record/825946}.}.


\textbf{Aggregation Layer}: We apply two matching mechanisms to aggregate the refined sentence representations, which are directly aimed at extracting relationships between the premise and the hypothesis. Concretely, we concatenate the representations of the premise $\bar{\bm{h}}_P^{\prime}$ and hypothesis $\bar{\bm{h}}_H^{\prime}$ in addition to their element-wise product ($\odot$) and the absolute value ($ | \cdot | $) of their difference, obtaining the vector $\bm{r}$. These last two operations, first proposed by \newcite{mou2015discriminative}, can be seen as a sentence matching strategy.

\begin{align}
	\bm{h}_{mul} &= \bar{\bm{h}}_P^{\prime} \odot \bar{\bm{h}}_H^{\prime}\\
    \bm{h}_{dif} &= | \bar{\bm{h}}_P^{\prime} - \bar{\bm{h}}_H^{\prime} | \\
    \bm{r}       &= [\bar{\bm{h}}_P^{\prime} ; \bar{\bm{h}}_H^{\prime} ; \bm{h}_{mul} ; \bm{h}_{dif}]
\end{align}



\textbf{Dense Layer}: Finally, $\bm{r}$ is fed to a fully-connected layer whose output is a vector containing the logits for each class, which are then fed to a softmax function for obtaining their probability distribution. The class with the highest probability is chosen as the predicted relationship between premise and hypothesis.

\section{Experiments}

To make our results comparable to the baselines reported in the Kaggle platform we randomly sampled 15\% of the SNLI corpus \cite{bowman2015snli} and added it to the MultiNLI corpus.

We used the pre-trained $300$-dimensional GloVe vectors trained on $840$B tokens \cite{pennington2014glove}. These embeddings were not fine-tuned during training and unknown word vectors were initialized by randomly sampling from the uniform distribution in $(-0.05, 0.05)$.

Each character embedding was initialized as a $20$-dimensional vector and the character-level LSTM output dimension was set to $50$. The word-level LSTM output dimension was set to $300$, which means that after concatenating word-level and character-level representations the word vectors for each direction are $350$-dimensional (i.e., $\bm{h}_i \in \mathbb{R}^{700}$).

For the Inner Attention Layer we defined the parameter $W$ as a square matrix matching the dimension of the concatenated vector $[\bar{\bm{h}}; \bm{h}_i]$ (i.e., $W \in \mathbb{R}^{1400\times 1400}$), and $\bm{v}$ as a vector matching the same dimension (i.e., $\bm{v} \in \mathbb{R}^{1400}$).  Both $W$ and $\bm{v}$ were initialized by randomly sampling from the uniform distribution on the interval $(-0.005, 0.005)$.

The final layer was created as a 3-layer MLP with 2000 hidden units each, and with ReLU activations.

Additionally, we used the Rmsprop optimizer with a learning rate of $0.001$. We applied dropout of $0.25$ only between the MLP layers of the Dense Layer.

Further, we found out that normalizing the capitalization of words by making all characters lowercase, and transforming numbers into a specific numeric token improved the model's performance while reducing the size of the embedding matrix. We also ignored the sentence pairs with a premise longer than $200$ words during training (for improved memory stability), and those without a valid label (``-'') both during training and validation.

Since one of the most conceptually important parts of our model was the raw sentence representation created in the Pooling Layer, we used four different methods for generating it (eqs. \ref{mean-pool} -- \ref{max-pool}). Results are reported in Table \ref{pool-results-table}.

We also tried using other architectures that rely on some sort of ``inner'' attention such as the \emph{self-attentive} model proposed by \newcite{lin2017selfattentive} and the \emph{co-attentive} model by \newcite{xiong2016dynamic}, but our preliminary results were not promising so we did not invest in fine-tuning them.

All the experiments were repeated without using character-level embeddings (i.e., $\bm{h}_i \in \mathbb{R}^{600}$).


\section{Results}

Table \ref{pool-results-table} presents the results of using different pooling strategies for generating a raw sentence representation vector from the word vectors. We can observe that that both the \emph{mean} method, and picking the last hidden state for both directions performed slightly better than the two other strategies, however at 95\% confidence we cannot assert that any of these methods is statistically different from one another.

This could be interpreted as if any of the four methods was good enough for capturing the overall meaning of the sentence, and the heavy lifting was done by the attention mechanism. It would be interesting to test these four strategies without the presence of attention to see whether it really plays an important role in this task or whether the predictive power lies within the sentence matching mechanism.


\begin{table}[htb]
\centering
\begin{tabular}{lcc}
\textbf{Method}      & \textbf{w/o. chars}     &\textbf{w. chars}     \\ \hline
\hline
\textit{mean}          	& 71.3 $\pm$ 1.2	  & 71.3 $\pm$ 0.7	         \\ \hline
\textit{sum}       	    & 70.7 $\pm$ 1.0	  & 70.9 $\pm$ 0.8	         \\ \hline
\textit{last}           & 70.9 $\pm$ 0.6	  & 71.0 $\pm$ 1.2	         \\ \hline
\textit{max}        	& 70.6 $\pm$ 1.1	  & 71.0 $\pm$ 1.1	         \\ \hline
\hline
\end{tabular}

\caption{Mean matched validation accuracies (\%) broken down by type of pooling method and presence or absence of character embeddings. Confidence intervals are calculated at 95\% confidence over 10 runs for each method.}
\label{pool-results-table}
\end{table}

\begin{table}[htb]
\centering
\begin{tabular}{lcc}
\textbf{Method}      & \textbf{w/o. chars}     &\textbf{w. chars}     \\ \hline
\hline
\textit{mean}          	& \textbf{72.3} 	  & 71.8 	         \\ \hline
\textit{sum}       	    & 71.6 	  & 71.6 	         \\ \hline
\textit{last}           & 71.4 	  & \textbf{72.1} 	         \\ \hline
\textit{max}        	& 71.1 	  & 71.6 	         \\ \hline
\hline
\end{tabular}

\caption{Best matched validation accuracies (\%) obtained by each pooling method in presence and absence of character embeddings.}
\label{pool-best-results-table}
\end{table}

Another interesting result, as shown by Table \ref{pool-results-table} and Table \ref{pool-best-results-table}, is that the model seemed to be insensitive to the usage of character embeddings, which was surprising because in our experiments with more complex models relying on shared information between premise and hypothesis, such as the one presented by \newcite{wang2017multiperspective}, the usage of character embeddings had a considerable impact in model performance\footnote{This type of models were not allowed in this competition which is why we do not report further on them.}.

In Table \ref{results-table} we report the accuracies obtained by our best model in both matched (first 5 genres) and mismatched (last 5 genres) development sets. We can observe that our implementation performed like ESIM overall, however ESIM relies on an attention mechanism that has access to both premise and hypothesis \cite{chen2017enhanced}, while our model's treats each one separately. This supports the notion that inner attention is a powerful concept.


\begin{table}[htb]
\centering
\resizebox{\columnwidth}{!}{%
\begin{tabular}{lccc}
\textbf{Genre}      & \textbf{CBOW}     &\textbf{ESIM}     &\textbf{InnerAtt} \\ \hline
\hline
Fiction          	& 67.5	  & 73.0	& 73.2         \\ \hline
Government       	& 67.5	  & 74.8	& 75.2         \\ \hline
Slate            	& 60.6	  & 67.9	& 67.2         \\ \hline
Telephone        	& 63.7	  & 72.2	& 73.0         \\ \hline
Travel           	& 64.6	  & 73.7	& 72.8         \\ \hline
\hline
9/11               & 63.2	  & 71.9	& 70.5         \\ \hline
Face-to-face       & 66.3	  & 71.2	& 74.5         \\ \hline
Letters            & 68.3	  & 74.7	& 75.4         \\ \hline
Oup                & 62.8	  & 71.7	& 71.5         \\ \hline
Verbatim           & 62.7	  & 71.9	& 69.5         \\ \hline
\hline
\textbf{MultiNLI Overall} 	 &\textbf{64.7}         	 &\textbf{72.2}           	 &\textbf{72.3}              \\ \hline
\end{tabular}
}
\caption{Validation accuracies (\%) for our best model broken down by genre. Both CBOW and ESIM results are reported as in \cite{williams2017broad}.}
\label{results-table}
\end{table}

We picked the best model based on the best validation accuracy score obtained on the matched development set ($72.257\%$). This model is as described in the previous section but without using character embeddings\footnote{Without the use of character embeddings, the sentence representations are 600-dimensional.}.

In addition, we created an ensemble by training 4 models as described earlier but initialized with different random seeds. The prediction is made by averaging the probability distributions returned by each model and then picking the class with the highest probability for each example. This improved our best test results, as reported by Kaggle, from $72.057\%$ to $72.247\%$ in the matched evaluation track, and from $72.055\%$ to $72.827\%$ in the mismatched evaluation track.

\section{Conclusions and  Future work}

We presented the model used by the team Rivercorners in the 2017 RepEval shared task. Despite being conceptually simple and not relying on shared information between premise and hypothesis for encoding each sentence, nor on tree structures, our implementation achieved results as good as the ESIM model.

As future work we plan to incorporate part-of-speech embeddings to our implementation and concatenate them at the same level as we did with the character embeddings. We also plan to use pretrained character embeddings to see whether they have any positive impact on performance.

Additionally, we think we could obtain better results by fine-tuning some hyperparameters such as the character embedding dimensions, the character-level LSTM encoder output dimension, and the Dense Layer architecture.


Further, we would like to see how different types of attention affect the overall performance. For this implementation we used the \emph{concat} scoring scheme (eq. \ref{att-scoring}), as described by \newcite{luong_effective_2015}, but there are several others that could provide better results.

Finally, we would like to exploit the structured nature of dependency parse trees by means of recursive neural networks \cite{tai2015improved} to enrich our initial sentence representations.

\section{Resources}
The code for replicating the results presented in this paper is available in the following link: \url{https://github.com/jabalazs/repeval_rivercorners}.

\section{Acknowledgements}
We thank the anonymous reviewers for helping us improve this paper through their feedback.


\newpage

\bibliography{emnlp2017}
\bibliographystyle{emnlp_natbib}

\end{document}